\documentclass[journal]{IEEEtran}
\usepackage{algorithm}
\usepackage{algorithmicx}
\usepackage{algpseudocode}
\usepackage{amsmath}
\usepackage{mathrsfs}
\usepackage{amsfonts}
\usepackage[numbers,sort&compress]{natbib}
\usepackage{amssymb}
\usepackage{bm}
\usepackage{enumerate}
\usepackage[pdftex]{graphicx}
\usepackage{lineno}
\usepackage{url}
\usepackage{color}
\usepackage{cases}
\emergencystretch=\maxdimen
\ifCLASSINFOpdf
\else
\fi

\begin{document}
%
\title{Concurrently Extrapolating and Interpolating Networks for Continuous Model Generation}
%
%
%

\author{Lijun~Zhao,
        Jinjing~Zhang,
        Fan~Zhang,
        Anhong~Wang,~\IEEEmembership{Member,~IEEE,}
        Huihui~Bai,~\IEEEmembership{Member,~IEEE,}
        and Yao~Zhao,~\IEEEmembership{Senior~Member,~IEEE}

\thanks{Corresponding author: Lijun Zhao. This work was supported in part by Doctoral Scientific Research Starting Foundation of Taiyuan University of Science and Technology(No. 20192023), Funding Awards for Outstanding Doctors Volunteering to Work in Shanxi Province(No. 20192055), National Natural Science Foundation of China(No. 61672373) and Key Innovation Team of Shanxi 1331 Project (KITSX1331).}
\thanks{L. Zhao and A. Wang are with the Institute of Digital Media \& Communication, Taiyuan University of Science and Technology, Taiyuan, 030024, P. R. China, e-mail: leejun@tyust.edu.cn}
\thanks{J. Zhang is with the Data Science and Technology, North University of China, Taiyuan, 030024, P. R. China, e-mail: leejun@tyust.edu.cn}
\thanks{F. Zhang is with School of Information and Communication Engineering, Beijing University of Posts and Telecommunications, Beijing, China, e-mail: zhangfan2015@bupt.edu.cn}
\thanks{H. Bai and Y. Zhao are with the Beijing Key Laboratory of Advanced Information Science and Network Technology, Institute of Information Science, Beijing Jiaotong University, Beijing, 100044, P. R. China, e-mail: yzhao@bjtu.edu.cn.}
}

%
%

\markboth{Journal of \LaTeX\ Class Files}
{Shell \MakeLowercase{\textit{et al.}}: Bare Demo of IEEEtran.cls for IEEE Journals}
%



\maketitle

\begin{abstract}
Most deep image smoothing operators are always trained repetitively when different explicit structure-texture pairs are employed as label images for each algorithm configured with different parameters. This kind of training strategy often takes a long time and spends equipment resources in a costly manner. To address this challenging issue, we generalize continuous network interpolation as a more powerful model generation tool, and then propose a simple yet effective model generation strategy to form a sequence of models that only requires a set of specific-effect label images. To precisely learn image smoothing operators, we present a double-state aggregation (DSA) module, which can be easily inserted into most of current network architecture. Based on this module, we design a double-state aggregation neural network structure with a local feature aggregation block and a nonlocal feature aggregation block to obtain operators with large expression capacity. Through the evaluation of many objective and visual experimental results, we show that the proposed method is capable of producing a series of continuous models and achieves better performance than that of several state-of-the-art methods for image smoothing.

\end{abstract}


\IEEEpeerreviewmaketitle
\section{Introduction}
\IEEEPARstart{T} {o} intelligently analyze image content and precisely identify scene objects, the image boundary is employed to provide many fundamentally vital clues that can generally be obtained via edge detection techniques \cite{Cei1, Cei2, Cei3, Cei4}. However, directly using these techniques always enable the detected edges to acquire some small yet unimportant discontinuity with strong gradients, which highly affects the performance of high-level real-world applications when using edge detection to extract low-level structural features. To resolve this problem, image structure extraction, namely, image smoothing, which is achieved by eliminating repeated texture elements, is usually treated as a prefiltering operation ahead of edge detection.

To the best of our knowledge, image smoothing, which is also called texture removal, has very wide application prospects in the fields of  image processing, computer graphics, pattern recognition, computational vision, etc \cite{Cei5, Cei6, Cei7}. When we wish to simultaneously conduct the image smoothing-oriented task and the texture removal-oriented task at the same time, these tasks can be considered as a special case of image decomposition. Decomposing undesirable textures from salient structures is a challenging and fundamentally ill-posed issue \cite{Cei8, Cei9, Cei10, Cei11, Cei12, Cei13, Cei14, Cei15, Cei16, Cei17, Cei18, Cei19, Cei20} since there is no clear definition of textures and structures. In fact, the prominent structures may consist of small textures on a certain scale, when viewed from a distance, but part of these small textures may possibly becomes remarkable structures when we observe them up close. Consequently, it is a hard problem to clearly identify them during image smoothing. The key challenges of texture removal or image smoothing issues can be summarized from several aspects as follows: (i) Distinguishing texture and structure is fundamentally ill-posed since there is no specific definition of them; (ii) Obtaining ground-truth labels with human-perceptual scores  is a time-consuming and laborious task, which leads to a limited number of available training data; (iii) Domain gaps exists between the synthetic (made by the texture and cartoon-like image) and real images datasets, when implicitly differentiating salient structures and fine details/textures; and (iv) When viewing the edge information of the same region at different scales, different people may regard them as textures or structures. Since these challenging problems still exist, we should deeply study these issues and make more contributions to the topic of image smoothing in theory and practice.

Classical convex optimization facilitates many classical approaches capable of achieving some excellent performances for image smoothing, but their algorithm complexity tends to be extremely high for optimal solutions, especially in an iterative way \cite{Cei5, Cei6, Cei7, Cei8}. Several recent works \cite{Cei1, Cei4, Cei14, Cei27, ICLR2016, resnet2, resnet1} have witnessed that the great progress is being made by deep learning techniques for low-level and high-level computer tasks. However, the learning of a deep neural network system depends entirely on the specific input and output labels; therefore, system can only obtain filtered images with a specific effect. When learning image smoothing operators, it is practically required that each learned model can generate various filtering effects of different magnitudes. However, most of existing deep learning approaches are not capable of achieving this functionality.

Although all of the problems discussed above for image smoothing require deep study, there are several fundamental yet key issues that require more urgent solution for most of the CNN-based approaches for practical applications, including the structure design of the deep image smoothing network with a strong capacity of image mapping, a training strategy, continuous model generation, etc. In this paper, we mainly study the problems of continuous network interpolation as well as learning multiple image smoothing operators without training multiple-times. Our contributions are summarized as follows:
\begin{itemize}
\item[a)] We generalize neural network interpolation as a more powerful model generation tool by tuning network weighting parameters. This tool facilitates the option of choosing a certain desired effect.
\item[b)] A simple yet effective model generation strategy is proposed to form a sequence of models with a single training, which only requires a set of specific-effect label images.
\item[c)] We propose a double-state aggregation (DSA) module to fuse the information of different stages or states, which can be easily inserted into most of the current neural network architectures.
\item[d)] We design a DSA image decomposition network with a local feature aggregation block and a nonlocal feature aggregation block to obtain deep image operators with large expression capacity by optimization learning.
\item[e)] Through many objective and visual quality comparisons, it is demonstrated that the proposed method achieves better performance than that many state-of-the-art image smoothing methods.
\end{itemize}

The rest of this paper is organized as follows: 1) The related works are given in Section II, and the problem formulation is presented in Section III; 2) The proposed generalized multiple-model generation framework is introduced in Section IV; 3) We detail the structures of double-state aggregation module in Section V, after which we provide a detailed description of double-state aggregation neural network in Section VI; and 4) The experimental simulation is presented in Section VII, followed by the conclusion in Section VIII.

\section{Related works}
As discussed above, there are many thorny problems for image smoothing. To resolve these problems, we first would like to comprehensively look back to traditional image smoothing techniques and deep image smoothing operators. Since deep network interpolation (DNI) topic plays a prominent part in learning image smoothing operator for continuous model generations, we also thoroughly review many literatures about this topic.

\subsection{Traditional image smoothing}

To remove the low contrast edges and maintain remarkable boundaries, Xu et al. formulated the image smoothing issue as a global localization problem of important edges with a strategy of counting the number of major boundary pixels \cite{Cei5}, which can suppress weak-gradient details. However, the smoothing strategy of this method was intrinsically to remove small nonzero gradients, which unavoidably led to the retention of certain isolated pixels with strong gradient boundary information. Following this work, Cheng et al. proposed a new approximation algorithm \cite{Cei15} to minimize the L0 gradient for two tasks of image smoothing and surface smoothing \cite{Cei5}. Later, by collecting and observing 200 images with types of structure-plus-texture, Xu et al. first verified that an inherent feature difference exists between textural and structural regions, which were discriminated by a measure of relative total variation (RTV). According to this measure, a global objective function was built with the data term and RTV regularizer \cite{Cei6}, whose solution is obtained by an iterative numerical solver. Most recently, Guo et al. introduced a concept of relative structure to identify the nature of mutual structure considering inconsistent structure as well as flat structure \cite{Cei19}, which was formulated as a nonconvex optimization problem similar to the relative total variation \cite{Cei6}. At the same time, Li et al. presented an efficient guided image smoothing by soft clustering as a kind of catalyzer to promote smoothing, which considered the spatial dependency between neighboring pixels as done in bilateral filtering.

Although these methods can resolve the problem of image smoothing, they cannot separate the input image as a group of images having different scale structures. Since different scale structures provide the nonlocal clues for scene content, it is of great importance to represent image structures at scales. Observing that Gaussian filtering can remove small structure edges but blur large-scale structures, Zhang et al. employed Gaussian filtering as an engine to distinguish image structure scales by dynamically rolling guidance filtering (RGF) \cite{Cei7}. However, RGF has a severely intractable problem of accurately locating the object discontinuity position for scale-aware filtering at each scale. To tackle this issue, Zhao et al. leveraged a local activity measurement, that is, a clipped and normalized variance or standard deviation, to drive the relative total variation (RTV) for smoothing, namely, LAD-RTVs \cite{Cei8}, which can obtain multiscale representation images with better sharp boundary preservation than that of RGF. Borrowing some rules from the rolling guidance filter \cite{Cei7}, Ham et al. jointly used image structural information from the static image and dynamically updated the image for a wide variety of applications, including image decomposition, flash/nonflash denoising, and depth super-resolution, etc \cite{Cei18}.

\subsection{Deep image smoothing operators}
Most optimization-based approaches computed in an iteratively updating manner can hardly perform fast image filtering within a given time, whose parallel acceleration cannot be realized by using the graphics processing block (GPU) hardware. In contrast, DNN learning-based filtering methods are always able to be quickly run with the aid of the GPU. Xu et al. first proposed a learning system of boundary protection smoothing operators based on a DNN \cite{Cei9}. This system only needed the input and output of each operator to learn corresponding filtering models of various operators without the need of considering nonconvex optimization and the essential principle or theory of image smoothing.

Unlike two-stage image filtering \cite{Cei9}, Zhao et al. proposed to learn deep image filtering operators for simultaneous image smoothing and edge detection \cite{Cei14}. As in the literature \cite{Cei14}, since the image texture smoother and structural edge extractor helped each other achieve better performance than a single image processing task, Guo et al. jointly considered them following a principle of iteratively extracting salient edges and then removing the fine details based on the salient edges \cite{Cei17}. To carry out edge-preserving filtering in real time, Liu et al. treated the class of low-level vision issues as recursive image filtering \cite{Cei10}, whose network was built upon the deep convolutional neural network as well as recurrent neural networks. According to the key idea of rolling guidance filter, Li et al. cast the image structure-texture separation task as deep joint image filtering by feeding the output from the previous iteration as the input of the current iteration \cite{Cei11}. Similar to \cite{Cei11}, Pan et al. directly applied the trained models of depth image denoising to scale-aware filtering and removed the small-scale structural details of the input images according to the rolling guidance strategy \cite{Cei13}.

For general deep image smoothing operators, their neural network parameters should be trained using strong supervision with explicit structure-texture pairs as label images. This explicit supervision led to the problem of having only fixed-style smoothed images predicted by deep operators. For this problem, Kim et al. used a DNN to get a deep variational prior and inserted it into an iterative smoothing process by using a fast algorithm of alternating minimization \cite{Cei16}. For learning-based texture estimation, Lu et al. formed a large dataset by merging numerous natural texture images together with some clean yet structure-only images \cite{Cei21}. They developed a deep texture prediction network and a semantic structure prediction network to accurately distinguish the texture from the structure for structure-awareness preservation.

Since, in general, no public dataset was developed for an objective comparison of different algorithms, Zhu et al. established a new dataset to form a benchmark \cite{Cei12}. By using this dataset, a new class loss of the weighted root mean squared error and weighted mean absolute error was introduced to train a general model by measuring the distances between the predicted image and a series of smoothed image labels, which was selected by 14 volunteers from several algorithms with different parameter settings. However, only one deep image smoothing operator can be learned in \cite{Cei12} when we use this dataset , which is labeled by humans according to visual perception. Although the above methods can learn deep image operator by data-driven training, it is troublesome to train a model for the same algorithm with different hyperparameters every time.

\begin{figure*}[t]
\centering
\includegraphics[width=7in]{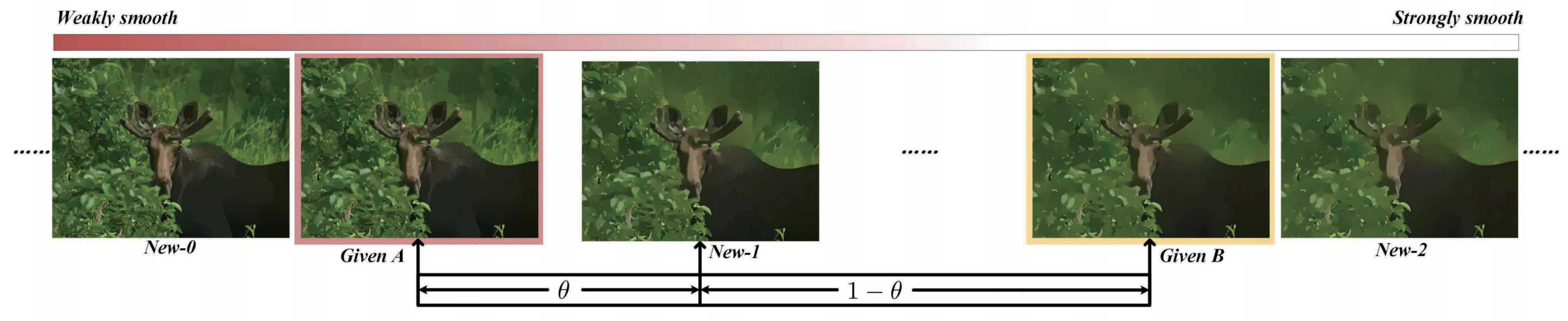}
\caption{The visualization of the smoothed image predicted with the models $New$-$\emph{0}$, $New$-$\emph{1}$ and $New$-$\emph{2}$ generated by concurrent extrapolation and interpolation (CEI) tool, when two models $A$ and $B$ are given ($\theta$ is a control parameter to adjust the smoothness of generated image).}
\label{int}
\end{figure*}

\subsection{Deep network interpolation}
In the early days, Upchurch et al. linearly interpolated pretrained deep convolutional features for automatic high-level semantic transformation \cite{Cei25}, which can be coarsely called a special kind of network interpolation. Later, Su et al. modified standard convolutions as a pixel-adaptive convolution (PAC) operation to spatially vary the convolution kernel for effectively learning guidance information \cite{Cei24}. After comparing the operations, it was found that they explore the harmonization of convolutional features, but they are two different tasks, among which the PAC is not designed for DNI.

To avoid repetitive training, Fan et al. first presented a mechanism of learning multiple deep parameterized image operators at the same time, which dynamically updated the deep base network's weights according to the weight learning network for learning multiple image operators \cite{Cei26}. To further reduce the computational cost, they also extended it to change the weights of only a single layer dynamically in the base network. In \cite{Cei28}, Kim et al. systematically reported CNN-based operator by carefully exploring its work principles. In the same period, Wang et al. gave a simple yet efficient strategy for DNI \cite{Cei23}, in which two or more correlated networks were linearly interpolated to smoothly control diverse imagery effects because almost all of the works targeted learning a deterministic mapping for the desired imagery effect. They widely apply DNI in super-resolution, image restoration, JPEG artifact removal and image-to-image translation as well as style transfer, etc.

Similarly, He et al. designed an adaptive feature modification, that is, AdaFM-Net, and it modified the channel-wise feature by adjusting an interpolation coefficient of a basic model and a modulation layer between a start and an end level \cite{Cei22}, since it is not easy to generalize deep neural network models toward continuously restoring contaminated images with unseen distortion-levels. The PAC and AdaFM-Net share some ideological similarities, for instance, both of them use a convolution layer to adapt it to a specific domain, but there are prominent differences between them. The former focuses on learning spatially adaptive guidance for deep joint image filtering, while the latter mainly studies a generalized model for continuously unseen distortion-level restoration without the need of an additional training stage \cite{Cei24}. Similar to AdaFM-Net, Wang et al. designed a CFSNet for image restoration, which controls latent convolutional features by adaptively learning coupling coefficients of diverse layers and channels \cite{Cei27}.

\begin{figure*}[t]
\centering
\includegraphics[width=6.3in]{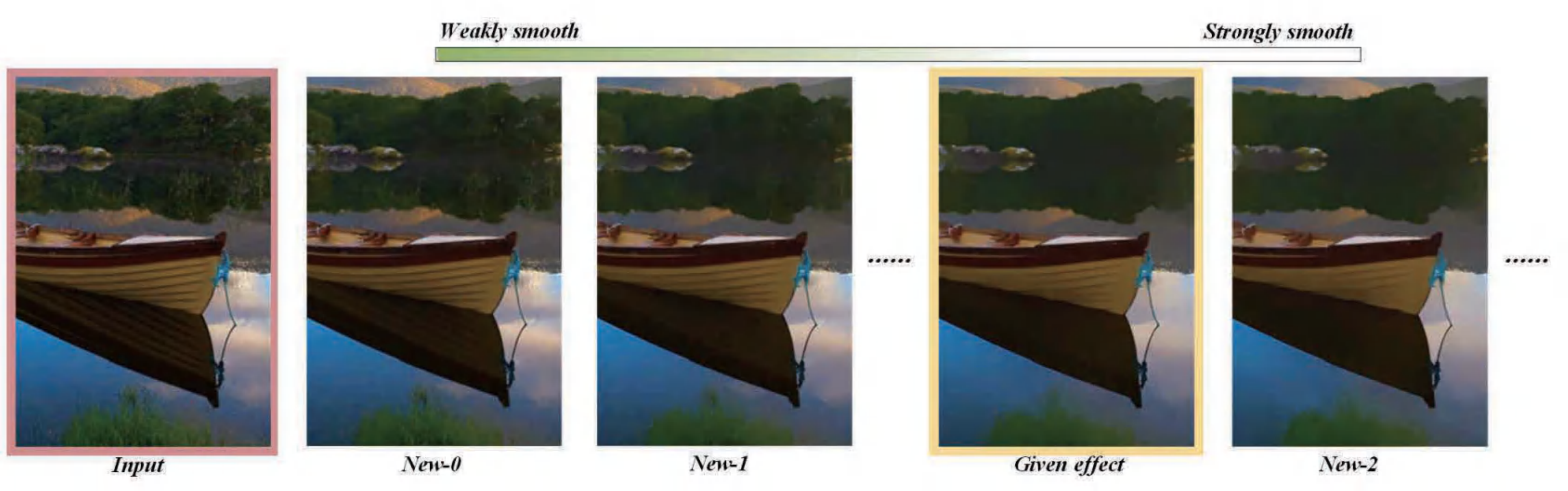}
\caption{The visualization of the smoothed image predicted with the models generated by concurrent extrapolation and interpolation (CEI) tool, when only a set of specific-effect label images is given.}
\label{onegiven}
\end{figure*}

\section{Problem formulation}
In theory, it is mathematically assumed that a captured image $I$ is able to be decomposed as a linear combination of a texture layer $T$ and a structure layer $S$, i.e., $I = T + S$. Given a natural image, an infinite number of solutions can be obtained according to a variety of priori constraints. In \cite{Cei12}, all the ground-truth smoothing results obtained by using many classically advanced image smoothing algorithms for training and testing are manually selected by 14 volunteers. According to human subjective perception, a high weight is assigned for each high-quality ground-truth smoothing image to form the quantitative measures \cite{Cei12}. Although a fixed-effect human-perceptual model is capable of being trained by using this dataset as the training one, there is no other available human-perceptual model with continuous imagery generation; therefore, the needs of different users cannot be met. Consequently, we should extend this fixed human-perceptual model to multiple models to satisfy various requirements. When directly training different models with different perceptual labels, we must have more ground-truth smoothing images with various assigned scores after these images are repeatedly watched and assessed by many volunteers. However, it is almost impossible to carefully and accurately label each image, especially for many datasets with millions of images, which is a labor-intensive and time-consuming process. Meanwhile, there is no efficient tool to quantify the human consistent assessment when continuous-perception scores are required.

To resolve these problems, we should fully take account of multiple model generation techniques such as the DNI, when putting forward an image smoothing approach based on artificial neural network. Given two correlated models based on a deep neural network, parameterized by various groups of parameters $V_{A}$ and $V_{B}$, a model generation of the DNI technique \cite{Cei23}, whose parameters are $V_{new}$, can be formulated as:
\begin{align}
V_{new} = \theta V_{A} + (1-\theta) V_{B}
\label{dni}
\end{align}
where $\theta$ is a weighting coefficient to control continuously specific degrees between imagery effect-$A$ and imagery effect-$B$, whose corresponding models are labeled as $A$ and $B$, e.g., image smoothness for texture removal. When any one model such as $A$ is trained with a group of specific-effect training datasets, the left model $B$ can be fine-tuned to adapt to a new group of specific-effect training datasets and vise versa.

\begin{figure*}[t]
\centering
\includegraphics[width=5in]{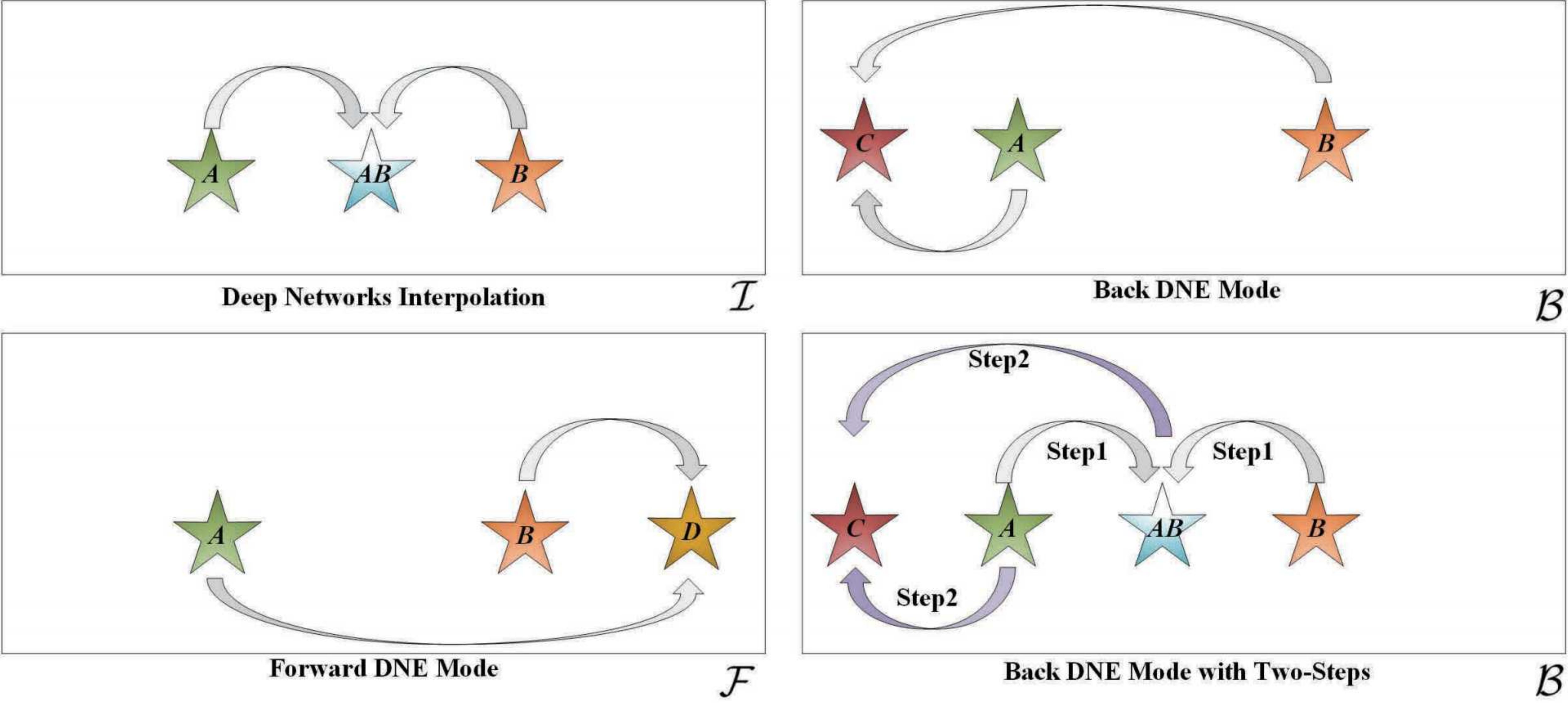}
\caption{The diagram of the generalized multiple model generation (Each pentagonal star represents a specific model using the same network architecture and $C$, $A$, $AB$, $B$ and $D$ are some examples of continuous modes).}
\label{final}
\end{figure*}

As exhibited above, the new model has the same network structure as the models $A$ and $B$, which has $\mathcal{L}$ layers. The intermediate-effect feature maps $I_{new}$ can be generated by linear parameter interpolation, when the feature maps $I_{new^{(i-1)}}$ in the $(i-1)$-th layer is fed into the $i$-th layer of the interpolated model. The network interpolation process can be represented as follows:
\begin{align}
I_{new^i} = I_{new^{(i-1)}}*V^{(i-1)}_{new}, i\in [0,...,\mathcal{L}]
\label{dnimage1}
\end{align}
\begin{align}
V^{i-1}_{new} = \theta V^{i-1}_{A} + (1-\theta) V^{i-1}_{B}
\label{dnimage2}
\end{align}
in which $*$ is the convolution operation. After simplification, we can obtain:
\begin{align}
I_{new^i} = \theta (I_{new^{(i-1)}}*V^{i-1}_{A})+(1-\theta) (I_{new^{(i-1)}}* V^{i-1}_{B})
\label{dnimage3}
\end{align}
This equation can be further rewritten as:
\begin{align}
I_{new^i} = \theta I^A_{new^{(i-1)}}+(1-\theta) I^B_{new^{(i-1)}}
\label{dnimage4}
\end{align}
where $I^A_{new^{(i-1)}}$ and $I^B_{new^{(i-1)}}$ are respectively, the predicted results from $I_{new^{(i-1)}}$ and $I_{new^{(i-1)}}$ using convolutional kernels of $V^{i-1}_{A}$ and $V^{i-1}_{B}$. From this equation, it is easy to understand that the intermediate-effect images can be obtained by the linear weighting method. When each layer is handled by the piecewise linear activation function, e.g., using the rectified linear unit (ReLU) function or its variants, such as the parametric ReLU and leaky ReLU function, to a certain extent, the interpolated network can maintain the characteristic of linearity for the output of each layer, which finally leads to continuous intermediate-effect images created by the new interpolated networks when continuously varying the trade-off parameter $\theta$. The majority of the above deep methods solely consider the study of deep network interpolation, and almost no work studies the topic of deep network extrapolation. Given only a set of specific-effect label images, there is no available technique for concurrent network interpolation and extrapolation. Thus, it is necessary to conduct a more in-depth study for continuous model generation.

\section{Proposed Generalized Multiple Model Generation Framework}

Currently, many image smoothing operators are always trained repetitively, when different explicit structure-texture pairs are employed as label images for each algorithm with different parameters. This kind of training often takes several days or
even two weeks, and it also consumes massive equipment resources. To resolve these challenging issues, we generalize the DNI technique as a more powerful model generation tool, namely, the concurrent extrapolation and interpolation (CEI) tool, to obtain more novel models. In other words, these models can produce a sequence of new images, in which the predicted effects are less than/more than the two given models, rather than only continuous imagery intermediate-effect transition, in comparison to the DNI technique. Our CEI tool refers to concurrent deep network extrapolation (DNE), which has a forward DNE mode {$\mathcal{F}$ and back DNE mode $\mathcal{B}$, and deep network interpolation $\mathcal{I}$ to generate continuous models, as shown in Fig. \ref{final}, whose predicted images are depicted in Fig. \ref{int} and can be written as:
\begin{eqnarray}
I^i_{new} = I^{(i-1)}_{new}*V^{(i-1)}_{{new}},
\label{ceinmage1}
\end{eqnarray}
\begin{eqnarray}
V^i_{new}=
\begin{cases}
  \frac{V^i_{A} - (1-\alpha) V^i_{B}}{1-(1-\alpha)}, &if (\mathcal{F}, 0 \le \alpha \le 1) \\
  \frac{V^i_{B} - \beta V^i_{A}}{1-\beta},         &if (\mathcal{B}, 0 \le \beta \le 1) \\
  \frac{\gamma V^i_{A} + (1-\gamma) V^i_{B}}{\gamma+(1-\gamma)}, &if (\mathcal{I}, 0 \le \gamma \le 1) \\
\end{cases}
\label{ceinmage2}
\end{eqnarray}

Given two models $A$ (less smooth) and $B$ (more smooth) of image smoothing as well as an extrapolating parameter $\alpha$ or $\beta$, for instance, we apply the forward DNE mode, which means that predicted images become smoother when $\alpha$ gradually increases. To clearly observe it, we can rewrite the forward DNE part of the above equation as $V^i_{new}=V^i_{B}-\frac{1}{\alpha}(V^i_{B}-V^i_{A})$, in which $\frac{1}{\alpha}$ is far greater than 1 since $\alpha$ is restricted between 0 and 1. As a result, the effect predicted by the generated model $V_{new}$, whose smoothness is denoted as $S_{B}$, is smoother than $V^i_{B}$ with a smoothness of $S_{A}$, that is, $S_{new}> S_{B}> S_{A}$. Similarly, we can obtain $S_{B}> S_{A}>S_{new}$ for back DNE mode. As described in the last section, $S_{B}> S_{new}> S_{A}$ when applying deep networks interpolation.

Since there is no further training of deep extrapolated networks from the given models, some problems such as color drift may appear in the predicted image, when directly using the above tool. To alleviate these problems, we first interpolate the intermediate-effect imagery with the generated model $AB$ between the imagery effect-$A$ from a given model $A$ and imagery effect-$B$ from a given model $B$. Since the similarity between the models $AB$ and $A$/$B$ is higher than that of the models $A$ and $B$, we use the models $AB$ and $A$/$B$ to form continuous models for deep network extrapolation. That is, as shown in Fig. \ref{final}, we can reformulate it as follows:
\begin{eqnarray}
I^i_{new} = I^{(i-1)}_{new}*V^{(i-1)}_{{new}},
\label{ceinmage3}
\end{eqnarray}
\begin{eqnarray}
V^i_{new}=
\begin{cases}
  \frac{(1+\alpha)V^i_{B}-(1-\alpha) V^i_{A}}{2\alpha}, &if (\mathcal{F}_{TS}, 0 \le \alpha \le 1) \\
  \frac{(1+\beta)V^i_{A}-(1-\beta) V^i_{B}}{2\beta}, &if (\mathcal{B}_{TS}, 0 \le \beta \le 1) \\
  \frac{\gamma V^i_{A} + (1-\gamma) V^i_{B}}{\gamma+(1-\gamma)}, &if (\mathcal{I}, 0 \le \gamma \le 1) \\
\end{cases}
\label{finalformula}
\end{eqnarray}
in which we propose a two-step (TS) deep network extrapolation to predict a series of images, as shown in Fig. \ref{int}. To infer this equation for the two-step deep network extrapolation, we should firstly obtain the intermediate model $AB$ via the deep network interpolation operation, after which the deep networks extrapolation operation is fulfilled to form extrapolated models. For our actual usage, we can directly use the formula in Eq. \ref{finalformula} rather than to generate extrapolated models step-by-step.

\begin{figure*}[t]
\centering
\includegraphics[width=5in]{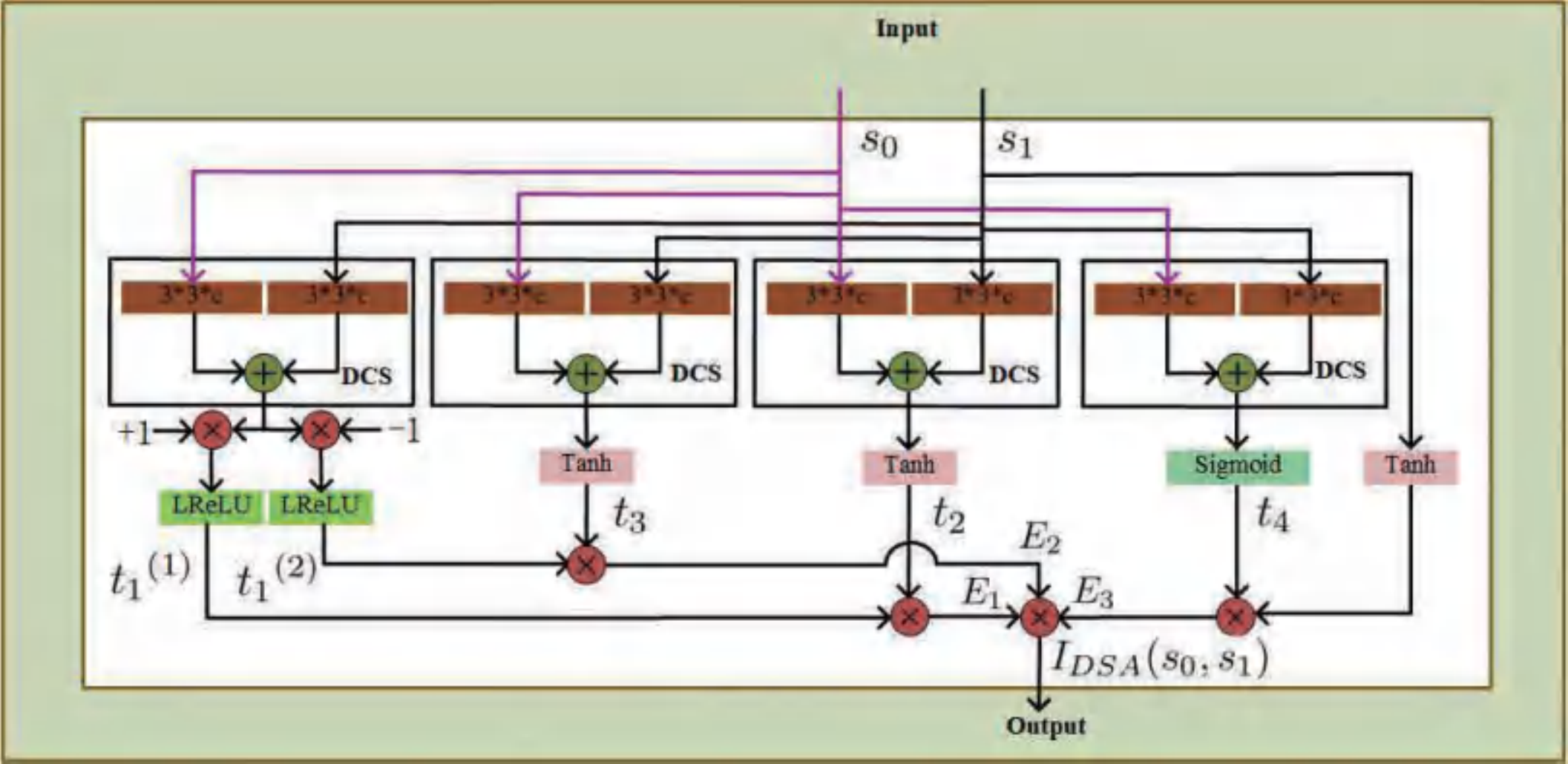}
\caption{The diagram of double-state aggregation module (DSA module, in which $s_0$, $s_1$ and $I_{DSA}$ are two inputs and the output of this module respectively).}
\label{DSAmodule}
\end{figure*}

This tool solely supports image smoothing operator learning to obtain two models from two given sets of datasets, so we cannot interpolate a series of new models directly when only a set of specific-effect label images is given. To be capable of learning multiple model generation operators for this case, we propose a simple yet effective model generation strategy to form a sequence of models, that is, mapping the input image back to the input image to continue training a new operator, after learning an operator of a set of specific-effect labels. Then, we can use these models to concurrently extrapolate and interpolate networks to obtain new models, whose predicted images are shown in Fig. \ref{onegiven}, toward continuous imagery transition for image smoothing.

\begin{figure*}[t]
\centering
\includegraphics[width=7in]{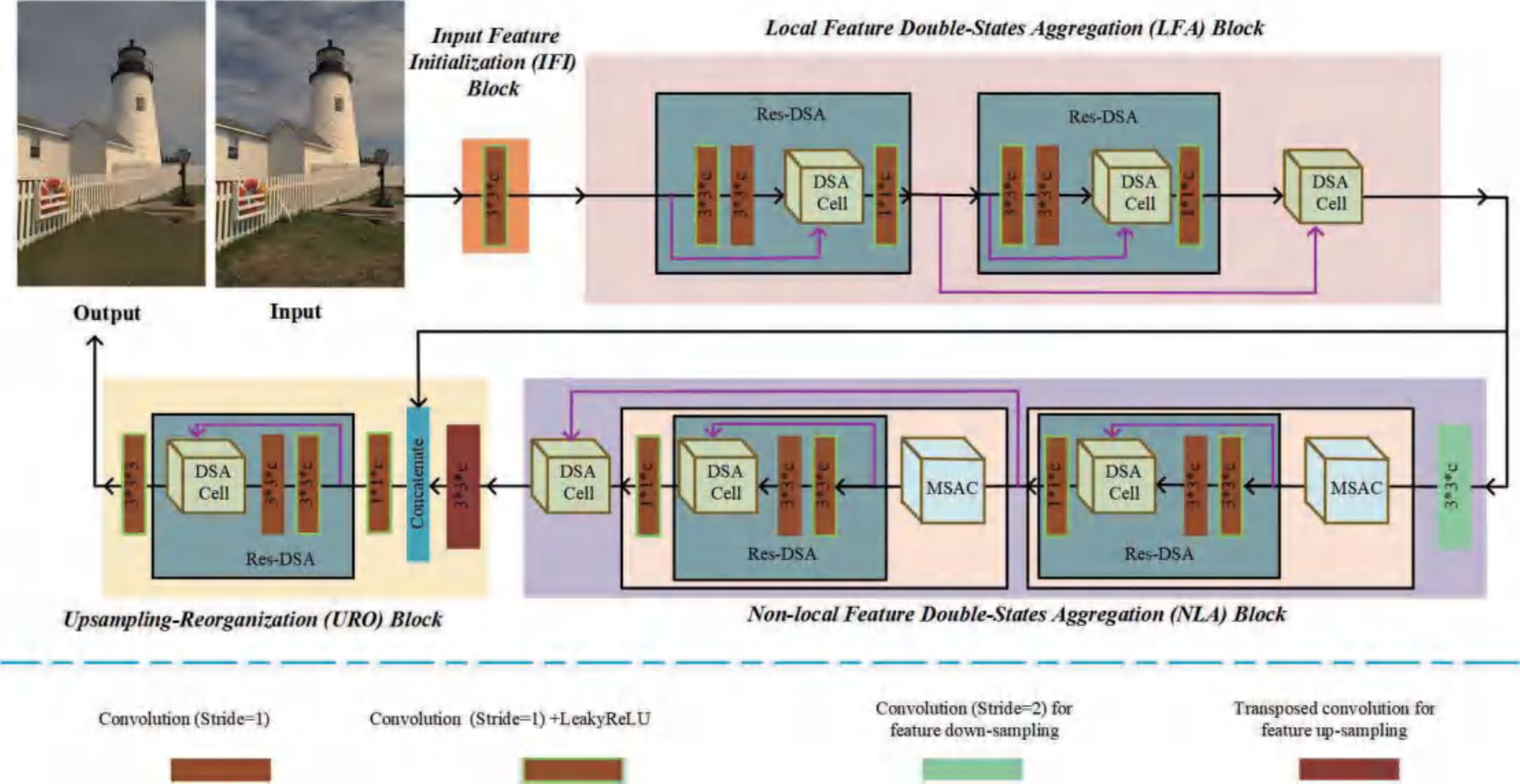}
\caption{The diagram for the proposed deep double-state aggregation neural network (DSAN).}
\label{smoothnet}
\end{figure*}

\section{Double-State aggregation (DSA) module}
In the past years, a general deep cascaded network has been generally built up by a sequence of convolution layers and activation functions \cite{cascadednetwork, itcsct}. However, since the derivative of each convolution layer will exponentially vary to be very large or very small, the gradients cannot adequately transmit from very deep layers to the shallow layers, and thus this cascaded network is not able to be well trained for task prediction when there is no use of expert knowledge tricks, including pretraining, layer-by-layer initialization, etc \cite{nature}. To relieve these issues, a series of network structures with a short-cut connection or residual connections such as ResNet and Inception-ResNet are explored to improve the accuracy of high-level and low-level computational vision task prediction \cite{resnet1, resnet2}. These networks always focus on the study of a better approach to gradient backpropagation and residual learning. However, they lose sight of the importance of the features fusion of different stages or different layers.

Although the recurrent neural network (RNN) and its variants such as long short-term memory (LSTM), and the gated recurrent neural network (GRU) have two inputs and two outputs, these recycled structures are designed for sequence data and are unsuitable for image data. To fully leverage diverse features at different stages or at various layers, we propose a double-state aggregation module, as shown in Fig. \ref{DSAmodule}. Different from RNN and its variants, our module structure modulates double states of inputs as multiple states and combines them together in a summation approach, after which these states are activated to obtain triple residual states. Finally, these triple residual states are summarized up to form a final fusion output for our DSA module.

Given two input states: $s_0$ and $s_1$, four pairs of dual convolution summation (DCS) units fuse these inputs together by the convolutional operation to permutate and combine these input features in the form of addition. Here, to avoid wasting the convolutional features with negative values, the outputs of the first DCS unit in Fig. \ref{DSAmodule} are multiplied by $-1$ and $+1$ before they are activated by the leaky ReLU function (LRelu, $\mathcal{R}$), that is,
\begin{align}
&t0=s_0*W_{1}^L+s_1*W_{1}^R+b1, \notag \\
&{t_1}^{(1)}=\mathcal{R}(+1 \odot t0), {t_1}^{(2)}=\mathcal{R}(-1 \odot t0)
\label{1dcs}
\end{align}
in which $W_{1}^L$, $W_{1}^R$ and $b1$ are two weights and the bias of the first DCS unit. After the activation of feature maps, they are respectively multiplied by tanh ($\mathcal{T}$) activated output features $t_2$ and $t_3$ of the second DCS unit and third DCS unit in an elementwise manner, and then we can obtain $E_1$ and $E_2$, that is,
\begin{align}
&E_1={t_1}^{(1)} \odot t_2,E_2={t_1}^{(2)} \odot t_3,\notag \\
&t_2=\mathcal{T}(s_0*W_{2}^L+s_1*W_{2}^R+b2),\notag \\
&t_3=\mathcal{T}(s_0*W_{3}^L+s_1*W_{3}^R+b3)
\label{23dcs}
\end{align}
in which $W_{2}^L/W_{3}^L$, $W_{2}^R/W_{3}^R$ and $b2/b3$ are two weights and the bias of the 2nd/3rd DCS unit. Additionally, $\odot$ denotes the elementwise multiplication. In the meantime, the outputs of the last DCS unit are activated by the sigmoid function $\mathcal{S}$, which is multiplied by tanh-activated input-b features in an elementwise manner. This operation can be written as:
\begin{align}
&E_3={t_4} \odot \mathcal{T}(s_1),\notag \\
&t_4=\mathcal{S}(s_0*W_{4}^L+s_1*W_{4}^R+b4)
\label{4dcs}
\end{align}
in which $W_{4}^L$, $W_{14}^R$ and $b4$ are two weights and the bias of the last DCS unit, respectively.
Finally, three kinds of features are combined utilizing addition, which can be written as:
\begin{eqnarray}
I_{DSA}(s_0, s_1) = E_1 + E_2 + E_3
\label{dcs}
\end{eqnarray}
in which $I_{DSA}$ is the final output of the $DSA$ module.

\section{Deep Double-State Aggregation Neural Network}
In this paper, we design a deep double-state aggregation neural network (DSAN) to learn image smoothing operators, as depicted in Fig. \ref{smoothnet}. It is composed of four parts: input feature initialization (IFI) block, a local feature double-state aggregation (LFA) block, a non-local feature double-state aggregation (NLA) block, and an upsampling-reorganization (URO) block. Since the IFI block is responsible for extracting $c$-channel feature maps from the input image $I$, only one convolutional layer with a spatial kernel size of $3\times3$ is used to obtain them, followed by the leaky ReLU function $\mathcal{R}$, which can be written as $ I_{IFI}=\mathcal{R}(I*w_{IFI})$, in which $w_{IFI}$ is the parameter of the IFI block. After feature initialization, we use a local feature aggregation block and a nonlocal feature aggregation block to obtain local and nonlocal features from distinct receptive field regions. As described above, we present a double-state aggregation (DSA) module, which can efficiently fuse features of different-stages or different-layers. Note that it can be easily inserted into most current network architectures. In the following, we will introduce the proposed LFA block, NLA block and URO block in detail.

\subsection{Local feature double-state aggregation (LFA) block}
Before introducing the LFA block, the ResNet-like structure with our DSA module is first defined as Res-DSA, which is marked as $\hbar$, as shown in Fig. \ref{smoothnet}. When it is the $i$-th times to use the structure of Res-DSA, this process is denoted as $\hbar_i$. In the Res-DSA, the short-cut connection and two consecutive operations of convolutional layers form two states as the inputs $s_0$ and $s_1$ of a DSA module. Different from ResNet, we use our DSA module to merge double-state together, rather than direct summation together. In the LFA, two Res-DSA are first cascaded together to extract local features, while both of them have various outputs in different stages of the LFA block, from which we can obtain two local states $\mathcal{S}_0=\hbar_1(I_{IFI})$ and $\mathcal{S}_1=\hbar_2(\mathcal{S}_0)$. Finally, these two states are merged by a new DSA module for different-stage information (DSI) aggregation, which is presented as:
\begin{eqnarray}
I_{LFA}=I_{DSA}(\mathcal{S}_0, \mathcal{S}_1)
\label{LFA}
\end{eqnarray}

\begin{figure}[t]
\centering
\includegraphics[width=3.3in]{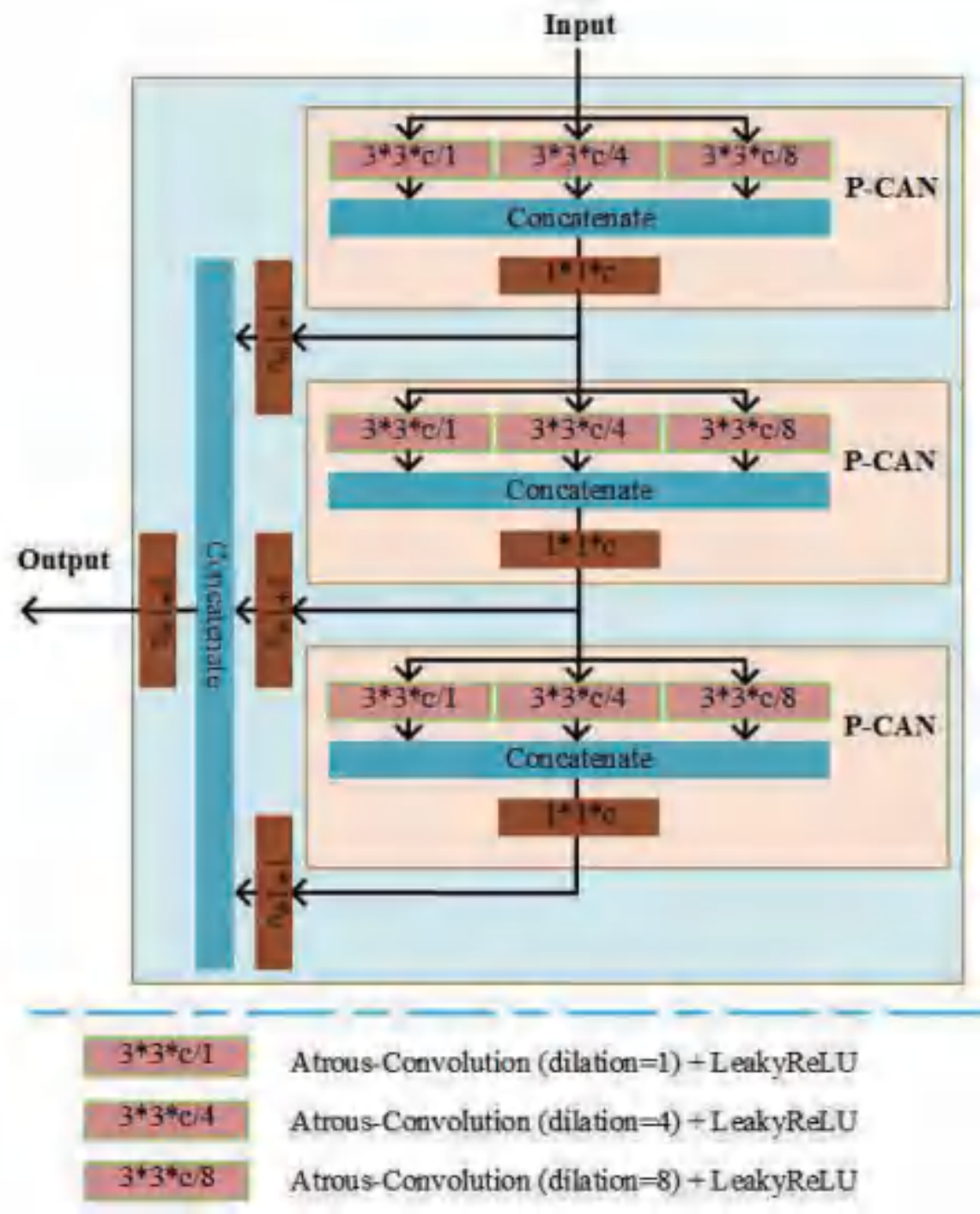}
\caption{The diagram of multiple stages atrous-convolution (MSAC).}
\label{MRAC}
\end{figure}

\subsection{Nonlocal feature double-state aggregation (NLA) block}
Since the receptive field of the LFA block is limited within certain local regions, we use multiple-stage atrous-convolution (MSAC) $\mathcal{M}$ to obtain nonlocal features, which is inspired by \cite{ICLR2016}. We denote this operation as $\mathcal{M}_i$ when it is the $i$-th time to use this network structure. Within the structure of MSAC, three parallel LReLU-activated atrous convolution layers with a spacing of 1, 4, and 8 between kernel elements is followed by a concatenation operation and a standard convolution layer with a spatial size of 1x1, namely, parallel connected atrous convolution (P-CAN), which can capture larger field features than that using several standard convolutions, as shown in Fig. \ref{MRAC}. To nonlocally perceive spatial correlations, we use three cascaded P-CAN to extract features. Before concatenating multiple-stage features from these three P-CAN together, we use three standard convolution layers with a spatial size of 1x1 to rearrange these features respectively. Finally, we use standard convolution layers to shrink the channel number after concatenation.

However, global feature extraction on the full-resolution feature maps with high computation complexities always takes very large memory. As a result, the spatial size of the input feature maps of $I_{LFA}$ should be diminished by a downsampling operation $\mathcal{D}$, that is, using a standard convolution layer with a stride of $2$, whose operation can be written as: $I^{\frac{1}{2}}_{LFA}=\mathcal{D}(I_{LFA})$. After that, we combine the MRAC together with Res-DSA as:
\begin{eqnarray}
I_{C}=\hbar_3(\mathcal{M}_1(I^{\frac{1}{2}}_{LFA}))
\label{LFA}
\end{eqnarray}
By combining MRAC and Res-DSA twice, two non-local states are formed as the inputs of a DSA module for DSI aggregation, that is,
\begin{eqnarray}
I_{NLA}=I_{DSA}(I_{C}, \hbar_4(\mathcal{M}_2(I_{C})))
\label{LFA}
\end{eqnarray}

\subsection{Upsampling-Reorganization (URO) block}

After local and nonlocal feature aggregation, outputs from both of the LFA block and the NLA block are concatenated and then fused by our URO block. Since the spatial size of the feature map $I_{NLA}$ is less than that of $I_{LFA}$, we cannot concatenate them along the channel dimension. Thus, the transposed convolutional operation $\mathcal{U}$ is used to upsample $I_{NLA}$ to full resolution, which can be written as: $I^F_{NLA}=\mathcal{U}(I_{NLA})$. After the concatenation of $I^F_{NLA}$ and $I_{LFA}$, we use a standard convolution $\mathcal{H}$ to halve the channel number of concatenated feature maps, $I_{H}=\mathcal{H}(I^F_{NLA},I_{LFA})$. To well reorganize these features $I_{H}$, we use the Res-DSA, which is followed by a convolution layer $\mathcal{C}$ to reconstruct a output $I_{URO}=\mathcal{C}(\hbar_5(I_{H}))$, which is the final predicted image $I_{DSAN}$.

\begin{table*}[t]
\centering
\caption{The setting of our entire model of DSAN (specific-effect B) and its a series of its variant models (Y=yes, N=no).}
\scriptsize
\begin{tabular}{cccccccc}
\hline
Operation/Model &      DSAN1 &      DSAN2 &      DSAN3 &      DSAN4 &      DSAN5 &      DSAN6 &       DSAN \\
\hline
      SSIM &          Y &          N &          Y &          Y &          Y &          Y &          Y \\

  Learning Strategy &          B &          B &          B &          B &          B &        A2B &      B2A2B \\

      MSAC &          N &          Y &          Y &          Y &          Y &          Y &          Y \\

   Res-DSA &          N &          Y &          N &          Y &          Y &          Y &          Y \\

   DSI aggregation &          N &          Y &          N &          N &          Y &          Y &          Y \\

\hline
\end{tabular}

\label{entireincomplete0}
\end{table*}

\begin{table*}[t]
\centering
\caption{The setting of our entire model of DSAN (specific-effect A)  and its a series of its variant models (Y=yes, N=no).}
\scriptsize
\begin{tabular}{cccccccc}
\hline
Operation/Model &      DSAN1 &      DSAN2 &      DSAN3 &      DSAN4 &      DSAN5 &      DSAN6 &       DSAN \\
\hline
      SSIM &          Y &          N &          Y &          Y &          Y &          Y &          Y \\

  Learning Strategy &        B2A &        B2A &        B2A &          A &          A &        B2A &      A2B2A \\

      MSAC &          N &          Y &          Y &          Y &          Y &          Y &          Y \\

   Res-DSA &          N &          Y &          N &          Y &          Y &          Y &          Y \\

   DSI aggregation &          N &          Y &          N &          N &          Y &          Y &          Y \\
\hline
\end{tabular}

\label{entireincomplete1}
\end{table*}

\begin{algorithm}[t]
\caption{Learning two deep image smoothing operators and generating continuous models with the proposed CEI tool}
\scriptsize
\begin{algorithmic}[1]
\renewcommand{\algorithmicrequire}{\textbf{Input:}}
\renewcommand{\algorithmicensure}{\textbf{Output:}}
\Require Given two sets of image pairs: $\Omega(\bm{I_A}, \bm{S})$ and $\Omega(\bm{I_B}, \bm{S})$; Given a network structure of our DSAN
\Ensure  Two specific-effect models and corresponding continuous interpolated or extrapolated continuous models: $M(A)$, $M(B)$ and $M(new)$;
\State Randomly initialize the weights of each convolutional layers in the network of DSAN for a specific effect $A$;
\State Optimize the loss function of Eq. (\ref{loss}) with $\Omega(\bm{I_A}, \bm{S})$ for a specific effect $A$ to get a start model of $M(A0)$;
\State Initialize the network for a specific effect $B$ using the parameters of $M(A0)$;
\State Optimize the loss function of Eq. (\ref{loss}) with $\Omega(\bm{I_B}, \bm{S})$ for a specific effect $B$ to get a model of $M(B)$;
\State Initialize the network for a specific effect $A$ using the parameters of $M(B)$;
\State Optimize the loss function of Eq. (\ref{loss}) with $\Omega(\bm{I_A}, \bm{S})$ for a specific effect $A$ to get a model of $M(A)$;
\State extrapolate and interpolate simultaneously for new continuous models using the proposed CEI tool according to Eq. (\ref{finalformula});
\State \textbf{return} $M(A)$, $M(B)$ and $M(new)$;
\end{algorithmic}
\end{algorithm}

\subsection{Learning strategy}
Given a pair of input images $I$ with a size of $m\times n$, our target is to learn a DSAN network for a specific image smoothing operator, e.g.,  to predict a specific effect-$A$ image $I_{A}$ using $Y_{A}$ as its label counterpart, whose loss function of training our framework can be formulated as:
\begin{align}
L(I, Y_{A}, I_{A})=\frac{\sum_{i\in\Omega} ||Y_{A}^i-I_{DSAN}^i||+ \varphi \times \varsigma(Y_{A}^i,I_{DSAN}^i)}{m\times n}
\label{loss}
\end{align}
in which $||\cdot||$ is the L1 norm, $\Omega$ is the pixel set of $I$, and $\varsigma(Y_{A}^i,I_{DSAN}^i)$ is the calculation of the structural similarity index (SSIM) of each pixel between $Y_{A}$ and $I_{DSAN}$. In addition, $\varphi$ is a predefined trade-off hyperparameter to harmonize a balance between L1-restricted data loss and SSIM loss when learning an image smoothing operator.

Recently, it has been proven that the predicted accuracy of fine-tuned models can be greatly improved \cite{icmew, wyh}. To learn two correlated image smoothing operators $A$ and $B$ using the same network structure, for example, when we have first obtained a CNN model $M(A0)$ for a specific effect $A$ by learning, all of the parameters of the operator $B$ can be initialized by the model of $M(A0)$ and are trained to obtain a model $M(B)$ of a specific effect $B$. Compared with $M(A0)$, the learned $M(B)$ is trained with a set of parameters as a pretrained model, and thus, a better model of the specific effect $A$ can be further trained to obtain a new model of $M(A0)$, when the parameters of $M(B)$ are used as the weight initialization of the network. Given two learned models acquired by sharing the same network structure, a group of continuous effect operators can be simultaneously extrapolated and interpolated using our CEI tool according to Eq. (\ref{finalformula}). To clearly see it, we summarize this procedure in \textbf{Algorithm-1}. For ease of expression, we denote the learning strategy in \textbf{Algorithm-1} as A2B2A. Similarly, we can use B2A to represent this strategy, when we first obtain a CNN model of $M(B0)$, which is followed by training a new model $M(A)$. If $M(B)$ is learned by using the parameters of $M(A)$ as its initialized counterpart after the learning of B2A, we refer it as B2AB.

\section{Experimental Simulation}
To verify the effectiveness of the proposed method, we give a qualitative and quantitative evaluation of image smoothing. The measurements of PSNR and SSIM are calculated as quality metrics to compare the performance of several comparative methods. Just as \cite{SSIMD}, the quantity of SSIM $\varsigma$ is transformed to decibels $\wp$ to make the quality factor legible according to $\wp=-10log_{10}(1-\varsigma)$ because the values of SSIM always tend to be close to each other.

\begin{figure*}[!htbp]
\centering
\includegraphics[width=5.6in]{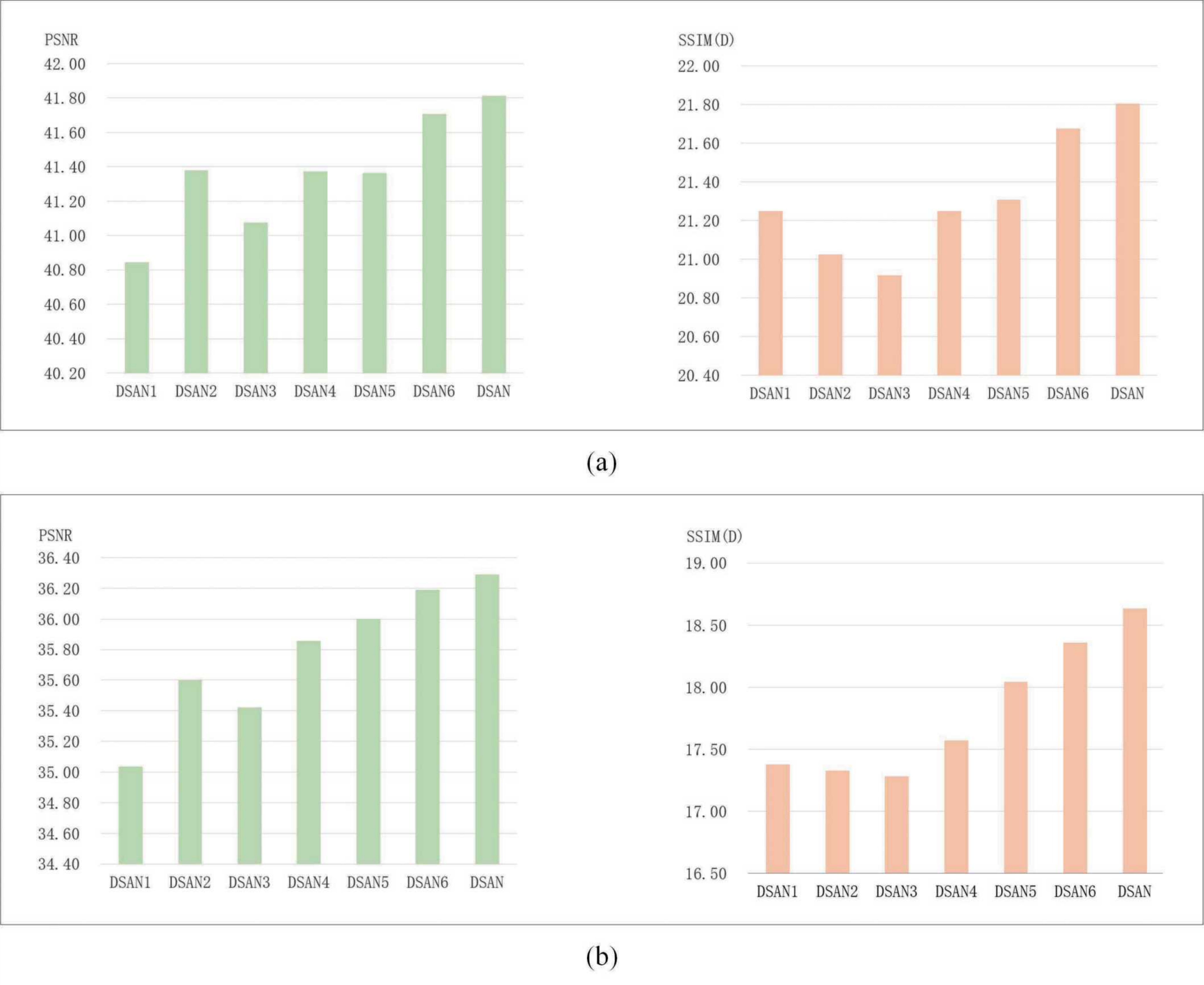}
\caption{The objective comparison between our entire model of DSAN and its a series of its variant models.}
\label{entire}
\end{figure*}

\subsection{Implementation Details}
To train our model, we use 400 images as the training dataset from \cite{Cei12}, of which 100 images of the testing images (Denoted as TIP100) are utilized to validate the efficiency of the proposed method. Meanwhile, 126 images are picked from the testing dataset of CUFED5 from \cite{cufed}. For simplification of testing, we resize the 226 testing images as 528*400, which can be found in this website{\footnote{https://github.com/mdcnn/Local-Activity-Driven-Filtering/}}. To augment the training data, we flip each image horizontally and vertically in a random style when training all the models. The proposed networks are trained using the Adam optimizer with $\beta_1$=0.1 and $\beta_2$=0.999. The initial learning rate is set to $2*10^{−4}$. The learning rate is decreased in a stepped-descent manner, with an attenuation factor of learning rate set as 0.5 and the learning rate decay period is set as 100. We implemented our deep model learning in the PyTorch framework and these models are trained using an NVIDIA GeForce GTX 2080TI GPU. It takes approximately ten hours to train our single DSAN model.

\subsection{Ablation Study}
To demonstrate the rationality of the network structure, we give a series of performance comparisons of our double-state aggregation neural network, when discarding some components and replacing them with the popular ResNet. Here, the specific-effect (A/B) of the L0 gradient minimization (denoted as L0GM) are learned, with a smoothness controlling parameter of $\lambda_{L0GM}$ set as $0.005/0.02$. At the same time, let $\kappa$ of the L0GM method to be 2, which is a recommended parameter to dominate the image boundary sharpness of natural images. To clearly observe and compare the divergences between our DSAN and its variants, TABLE. \ref{entireincomplete0} and TABLE. \ref{entireincomplete1} are provided. Note that the models of our DSAN and its variants retain similar numbers of network parameters.

The objective quality comparison of each model is given when testing on the TIP2019 dataset in Fig. \ref{entire}, from which it be found that, without the restriction of SSIM loss for training, both of SSIM(D) and PSNR of the entire DSAN model drop off up to 0.69dB/0.43dB and 1.3/0.78, compared to DSAN2, when using the L0GM method to generate label images with $\lambda_{L0GM}=0.02/0.005$. The performance of DSAN's variants such as DSAN1(B), DSAN2(B) and DSAN3(B) degrades when removing some components such as the MSAC, Res-DSA, and DSI aggregation within the network structure of the proposed DSAN. From objective comparisons between DSAN5 and DSAN (or between DSAN6 and DSAN), it is apparent that the performance of the proposed method can be greatly improved when using the proposed learning strategy described in \textbf{Algorithm-1}.

\begin{table}[t]
\centering
\caption{The objective quality comparison of image smoothing results predicted by different learned  operators({\bf{\color{red}COLOR}} has the best performance, {\bf{\color{cyan}COLOR}} is the second one, and {\bf{\color{blue}COLOR}} is the third one).}
\scriptsize
\begin{tabular}{c|ccccc}
\hline
\multicolumn{ 2}{c}{To-be-learned Models} & \multicolumn{ 2}{c}{L0GM(0.00431)} & \multicolumn{ 2}{c}{L0GM(0.02)} \\
\hline
   Dataset & Method/Measurement &       PSNR &    SSIM(D) &       PSNR &    SSIM(D) \\
\hline
\multicolumn{ 1}{c|}{} &       DEAF \cite{Cei9}  &     ---    &     ---    &     31.25  &     14.40  \\

\multicolumn{ 1}{c|}{} &     PR2019 \cite{Cei14} &     ---    &     ---    &     32.62  &     {\bf {\bf{\color{red}18.63} }}  \\

\multicolumn{ 1}{c|}{} &   PAMI2019 \cite{Cei26} &     38.26  &     17.77  &     {\bf{\color{blue}35.66}}  &     16.36  \\

\multicolumn{ 1}{c|}{TIP100} & VDCNN(TIP2019) \cite{Cei12} &     37.62  &     15.62  &     31.65  &     11.59  \\

\multicolumn{ 1}{c|}{} & ResNet(TIP2019) \cite{Cei12} &     {\bf{\color{blue}41.17}}  &     {\bf{\color{blue}21.19}}  &     35.05  &     {\bf{\color{blue}17.42}}  \\

\multicolumn{ 1}{c|}{} & DnCNN(19,256) \cite{zk} &     41.05  &     20.56  &     33.91  &     15.44  \\

\multicolumn{ 1}{c|}{} & DSAN(w/o SSIM loss) & {\bf {\color{cyan}41.94} } & {\bf {\color{cyan}21.55} } & {\bf {\color{cyan}36.07} } & {\bf {\color{cyan}18.21} } \\

\multicolumn{ 1}{c|}{} &       DSAN & {\bf {\color{red}42.37} } & {\bf {\color{red}22.08} } & {\bf {\color{red}36.29} } & {\bf {\color{red}18.63} } \\
\hline
\multicolumn{ 1}{c|}{} &       DEAF \cite{Cei9} &     ---    &     ---    &     31.22  &     13.85  \\

\multicolumn{ 1}{c|}{} &     PR2019 \cite{Cei14} &     ---    &     ---    &     31.60  &     14.99  \\

\multicolumn{ 1}{c|}{} &   PAMI2019 \cite{Cei26} &     39.13  &     17.45  &     {\bf{\color{blue}35.93}}  &     16.20  \\

\multicolumn{ 1}{c|}{CUFED5} & VDCNN(TIP2019) \cite{Cei12} &     37.25  &     13.68  &     31.34  &     10.42  \\

\multicolumn{ 1}{c|}{} & ResNet(TIP2019) \cite{Cei12} &     {\bf{\color{blue}41.39}}  &    {\bf{\color{blue} 19.79}}  &     35.00  &     {\bf{\color{blue}16.29}}  \\

\multicolumn{ 1}{c|}{} & DnCNN(19,256) \cite{zk} &     40.99  &     18.54  &     33.61  &     13.95  \\

\multicolumn{ 1}{c|}{} & DSAN(w/o SSIM loss) & {\bf {\color{cyan} 42.22} } & {\bf {\color{cyan} 20.41} } & {\bf {\color{cyan} 36.16} } & {\bf {\color{cyan} 17.06} } \\

\multicolumn{ 1}{c|}{} &       DSAN & {\bf {\color{red} 42.67} } & {\bf {\color{red} 20.60} } & {\bf {\color{red} 36.39} } & {\bf {\color{red} 17.21} } \\
\hline
\end{tabular}
\label{OBJJ}
\end{table}

\subsection{Quality Comparison}
To validate the efficiency of the proposed DSAN model for image smoothing, we compare our entire model of DSAN and DSAN(w/o SSIM loss) with several existing state-of-the-art approaches such as DEAF \cite{Cei9}, PR2019 \cite{Cei14}, PAMI2019 \cite{Cei26}, VDCNN(TIP2019) \cite{Cei12}, ResNet(TIP2019) \cite{Cei12} and DnCNN(19,256) \cite{zk} in terms of PSNR and SSIM(D), as displayed in TABLE. \ref{OBJJ}. In this table, the proposed method of DSAN(w/o SSIM loss) does not make use of SSIM loss during training in contrast to DSAN.

\begin{figure*}[!htbp]
\centering
\includegraphics[width=6.0in]{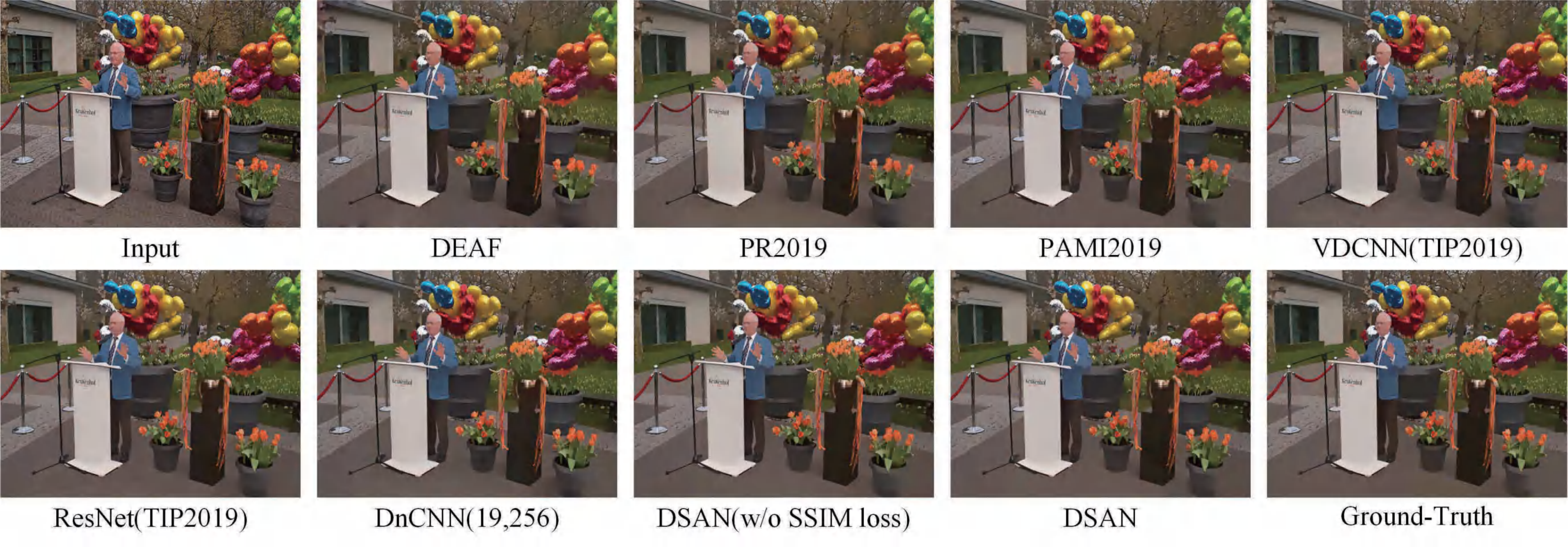}
\caption{The visual comparison of image smoothing results when using different network architectures.}
\label{sota}
\end{figure*}

TABLE. \ref{OBJJ} shows that our DSAN consistently has the highest objective quality measurements for image smoothing, while the objective quality of DSAN(w/o SSIM loss) ranks second, when comparing these latest approaches and DSAN(w/o SSIM loss). In other words, these objective results, to a certain extent, reflect that the design of the DSAN network architecture is rational and the proposed algorithm is effective in training. Meanwhile, the performance of ResNet(TIP2019) \cite{Cei12} on most occasions is better than that of DEAF \cite{Cei9}, PR2019 \cite{Cei14}, PAMI2019 \cite{Cei26}, VDCNN(TIP2019) \cite{Cei12}, and DnCNN(19,256) \cite{zk} since ResNet(TIP2019) \cite{Cei12} uses the residual convolution (Res-conv) with the skipping connection, which is conducive to the fast convergence of the network with the help of easy gradient backpropagation. Similar to ResNet(TIP2019), the residual reconstruction of VDCNN(TIP2019) \cite{Cei12} and DnCNN(19,256) \cite{zk} in a skip-connection manner only predicts the residual information rather than direct prediction with the cascaded convolutional network, but they have less capacity than does ResNet(TIP2019) with Res-conv as its fundamental building blocks since VDCNN(TIP2019) \cite{Cei12} and DnCNN(19,256) \cite{zk} only use the skip connection one time. Although the network of PAMI2019 \cite{Cei26} without using residual reconstruction is a cascaded neural network composed of three standard convolutional layers and seven Res-conv blocks that are followed by three standard convolutional layers, its predicted images are less similar to the ground-truth images than those of ResNet(TIP2019); however, PAMI2019 \cite{Cei26} achieves higher performance than that of VDCNN (TIP2019) \cite{Cei12} in the terms of PSNR and SSIM(D). In these methods, VDCNN(TIP2019) \cite{Cei12} has the worst performance on SSIM(D) measurements, that is, other methods can better preserve the edge structures. In addition, all of the other methods achieve higher objective quality than that of DEAF \cite{Cei9} when measuring the PSNR of the results predicted by these methods.

\begin{figure*}[!htbp]
\centering
\includegraphics[width=6.3in]{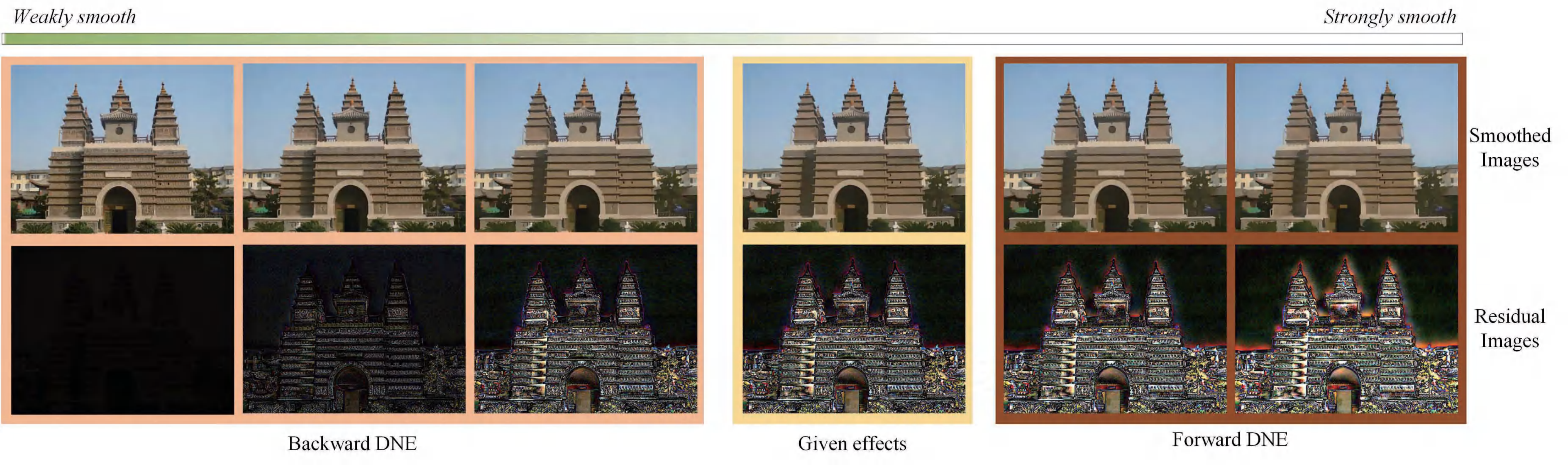}
\caption{The visualization of smoothed images (in the first row) predicted with models generated by our CEI tool and the residuals (in the second row) between these images and the input image, when only a set of specific-effect label images generated from \cite{Cei12} is given for training.}
\label{CM}
\end{figure*}
To more clearly see the performances of these methods, we also give a visual comparison of these approaches, as provided in Fig. \ref{sota}. From these figures, it can be seen that the image smoothing results of our DSAN(w/o SSIM loss) and our DSAN are the most similar to the ground truth compared with all of the other methods, while DSAN can better preserve image structural contours than DSAN(w/o SSIM loss). At the same time, the smoothness of the results predicted by DnCNN(19,256) \cite{zk} and VDCNN(TIP2019) \cite{Cei12} is less than that of all of the other comparative approaches since the architectures of DnCNN(19,256) \cite{zk} and VDCNN(TIP2019) \cite{Cei12} directly cascade convolutional layers one-by-one. Among these methods, the DEAF \cite{Cei9} is a special approach, which first uses a convolution neural network to estimate smoothed image gradients, and then these gradients are used to remove image textures by a traditional optimization technique. Apparently, the results of DEAF \cite{Cei9} are affected by both the estimated gradients and the corresponding optimization technique. From Fig. \ref{sota}, it can be easily found that the smoothed image predicted by DEAF \cite{Cei9} has some false-boundary artifacts around the image's outermost regions in Fig. \ref{sota}, which do not appear in the images derived from the other methods.

As discussed above, F. Zhu et al. provided a benchmark to learn both VDCNN and ResNet models with the help of subjectively perceptive weighted loss functions for edge-preserving image smoothing, in which seven classical image smoothing algorithms are used to produce the ground-truth label image. Although the trained models of VDCNN and ResNet from \cite{Cei12} are able to obtain some satisfactory results, they can only produce fixed-effect smoothed images, which cannot produce some similar results to meet the requirements of different users. To generate a series of smoothed images, we use their model-predicted images from the ResNet model of \cite{Cei12} as our labels to train the proposed network of DSAN. The visual results from continuous models generated by our model generation tool are shown in Fig. \ref{CM}, from which it is seen that the proposed framework can produce a series of continuous models by simultaneously extrapolating and interpolating neural networks for image smoothing. Compared to that of DNI \cite{Cei23}, the priority of the proposed model generation tool lies in the capability of our framework to create a group of new images with continuous effects when only the image dataset and its corresponding labeling images with specific effects are given. In this case, the original DNI fails to form multiple continuous models. It is noteworthy that our model generation tool can produce more models than DNI when concurrently extrapolating and interpolating networks. However, DNI can only obtain the intermediate-effect images, which is bound to restrict its wide practical application.

\section{Conclusion}
In this paper, we propose a powerful model generation tool by generalizing continuous network interpolation. Meanwhile, a simple yet effective model generation strategy is given to form a sequence of models when only a set of specific-effect label images is provided. In addition, we present a double-state aggregation (DSA) module to learn image smoothing operators, which can be easily inserted into most current network architectures. Based on this module, we propose a double-state aggregation neural network structure with large expression capacity to learn image smoothing operators. To validate the rationality of our network design, we conduct many experiments and provide the experimental results for our entire model of DSAN and its series of variant models. Numerous objective and visual experimental results show that the proposed method is better than several novel methods in terms of PSNR and SSIM. Note that the proposed method, which is not restricted to textural removal, has wide practical applications including image style transfer, image-to-image translation and so on.




\bibliographystyle{IEEEtran}
\bibliography{IEEEfull,CEIN}
\end{document}